\documentclass[lettersize,journal]{IEEEtran}
\usepackage{amsmath,amsfonts}
\usepackage{algorithmic}
\usepackage{algorithm}
\usepackage{array}
\usepackage[caption=false,font=normalsize,labelfont=sf,textfont=sf]{subfig}
\usepackage{textcomp}
\usepackage{stfloats}
\usepackage{url}
\usepackage{verbatim}
\usepackage{graphicx}
\usepackage{cite}
\usepackage{amsmath, bm}
\usepackage[figuresright]{rotating}
\usepackage{color}

\definecolor{red}{rgb}{1,0,0} 
\definecolor{blue}{rgb}{0,0,1} 
\definecolor{purple}{rgb}{0.5,0.0,0.2}

\usepackage{amsmath}
\usepackage{amssymb}
\usepackage{threeparttable}
\usepackage{algorithm}
\usepackage{algorithmic}
\usepackage{multirow}

\begin{document}

\title{Region-Enhanced Feature Learning for Scene Semantic Segmentation}

\author{Xin Kang, Chaoqun Wang, Xuejin Chen, ~\IEEEmembership{Member,~IEEE}
\thanks{Xin Kang and Xuejin Chen are with the National Engineering Laboratory for Brain-inspired Intelligence Technology and Application, University of Science and Technology of China, Hefei 230026, China (e-mail: kx111@mail.ustc.edu.cn, xjchen99@ustc.edu.cn). Chaoqun Wang is with South China Normal University, Guangzhou 510620, China (e-mail: cqwang96@gmail.com)}
\thanks{Xuejin Chen is the corresponding author.}
\thanks{This work is supported in part by the National Natural Science Foundation of China (NSFC) under Grant 62076230; in part by the Fundamental Research Funds for the Central Universities under Grant WK3490000008; and in part by Microsoft Research Asia.}
}



\maketitle

\begin{abstract}
Semantic segmentation in complex scenes relies not only on object appearance but also on object location and the surrounding environment.
Nonetheless, it is difficult to model long-range context in the format of pairwise point correlations due to the huge computational cost for large-scale point clouds.
In this paper, we propose using regions as the intermediate representation of point clouds instead of fine-grained points or voxels to reduce the computational burden. 
We introduce a novel \emph{Region-Enhanced Feature Learning Network} (REFL-Net) that leverages region correlations to enhance point feature learning.
We design a \emph{region-based feature enhancement} (RFE) module, which consists of a Semantic-Spatial Region Extraction stage and a Region Dependency Modeling stage.
In the first stage, the input points are grouped into a set of regions based on their semantic and spatial proximity.
In the second stage, we explore inter-region semantic and spatial relationships by employing a self-attention block on region features and then fuse point features with the region features to obtain more discriminative representations.
Our proposed RFE module is plug-and-play and can be integrated with common semantic segmentation backbones.
We conduct extensive experiments on ScanNetV2 and S3DIS datasets and evaluate our RFE module with different segmentation backbones.
Our REFL-Net achieves 1.8\% mIoU gain on ScanNetV2 and 1.7\% mIoU gain on S3DIS with negligible computational cost compared with backbone models. 
Both quantitative and qualitative results show the powerful long-range context modeling ability and strong generalization ability of our REFL-Net. 
\end{abstract}

\begin{IEEEkeywords}
semantic segmentation, region extraction, region dependency modeling.
\end{IEEEkeywords}

\section{Introduction} \label{section:Introduction}

\begin{figure}[!t]
  \centering
  \includegraphics[width=1.0\linewidth]{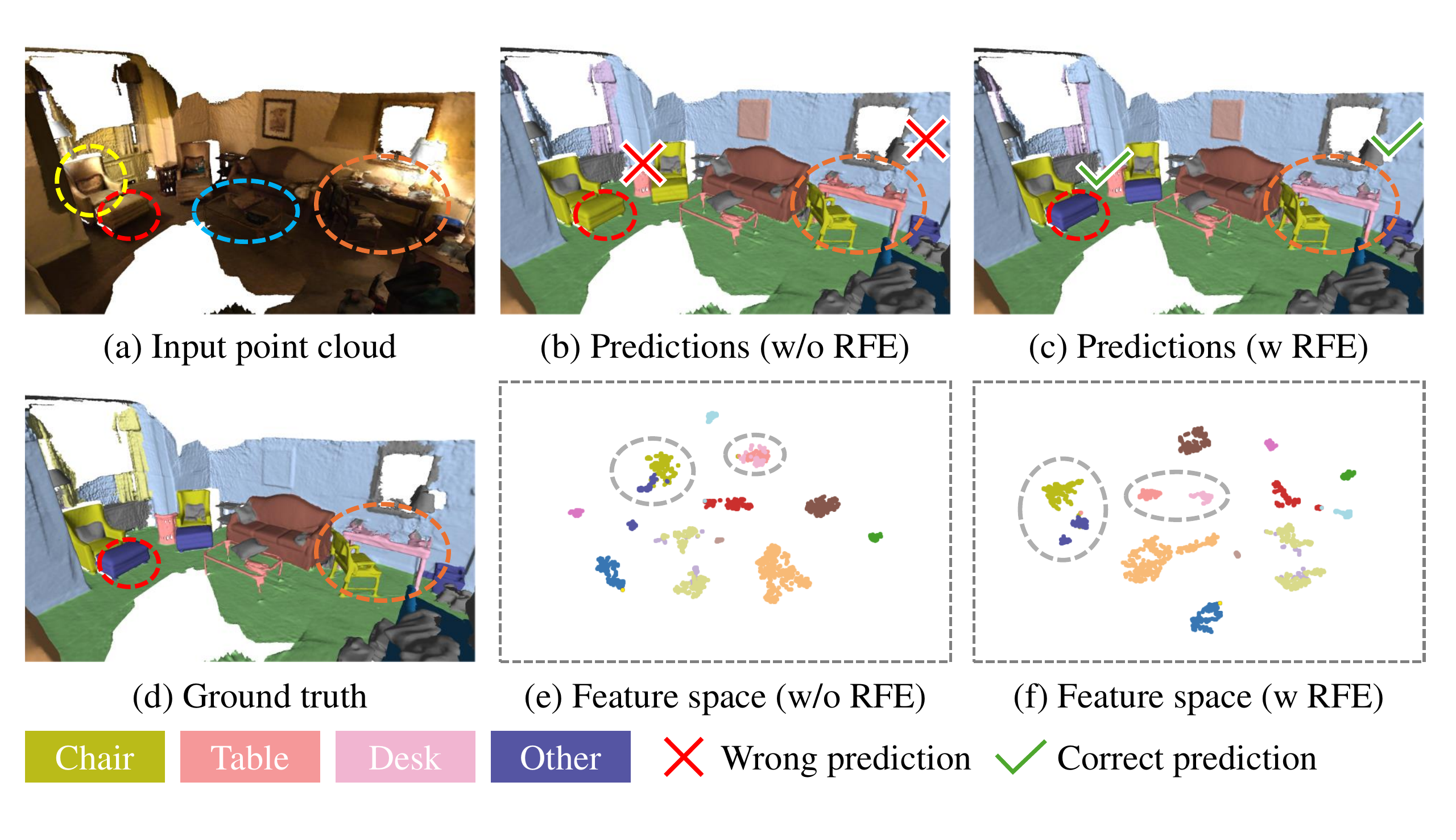}
  \vspace{-8mm}
  \caption{Object category confusion in semantic segmentation. In the point cloud (a), the footrest (red circle) belonging to \textbf{`Other'} category has a similar texture to the \textbf{`Chair'} (yellow circle), while \textbf{`Desk'} (orange circle) has a similar geometry to \textbf{`Table'} (blue circle). With only local feature aggregation, the baseline model predicts wrong categories for the footrest and the desk (b). With our region-based feature enhancement module, the model can well integrate long-range context and make correct category predictions (c). The feature distributions shown in (e) and (f) demonstrate that our method with RFE learns more distinctive point features.}
  \label{fig:confusion}
\vspace{-4mm}
\end{figure}

Point cloud semantic segmentation aims to classify each point into different semantic classes and plays an important role in scene understanding. It has been widely applied in autonomous driving, robotics, and augmented reality.
While appearance is a critical clue for identifying object categories, different objects may have similar local appearances, leading to confusing predictions, as shown in Fig.~\ref{fig:confusion} (a) and (b). This is mainly because different furniture objects in indoor scenes usually have similar materials, such as wood, fabric, etc.
Therefore, long-range context information, including global locations and the surrounding environment, is significantly important to resolve the classification ambiguity challenge. 

Most previous methods~\cite{qi2017pointnet++,wang2017cnn,graham20183d,li2018pointcnn,wu2019pointconv,choy20194d,yan2020pointasnl, qian2022pointnext} follow a down-sampling and up-sampling framework to progressively aggregate point features. These methods focus on designing effective local feature aggregation strategies from bottom to up, thus seldom exploring long-range dependencies across the whole scene.
Along another research line, attention-based networks \cite{engel2021point, guo2021pct, zhao2021point, lai2022stratified} aim to exploit long-range context in point clouds by employing self-attention on points. 
Since the high computational cost of self-attention layers scales with the number of queries and keys, these methods impose significant limitations on the input point number and receptive field size.
For instance, two object classification approaches \cite{engel2021point, guo2021pct} restrict the number of input points to 1,024 and 2,048, respectively, impairing their generalization ability for semantic segmentation in large-scale point clouds.
Stratified Transformer \cite{lai2022stratified} and Point Transformer\cite{zhao2021point} perform self-attention in local regions to handle large-scale point clouds, resulting in a limited receptive field.
Hence, capturing point-level pairwise correlations with limited computational resources remains a significant challenge.

In this paper, we propose to exploit long-range context by applying attention mechanisms at the region level instead of the point level to reduce the computational cost for large-scale point clouds. 
We introduce region as a compact intermediate representation, while each region stands for a group of points that are close both semantically and spatially.
Since the number of regions is much smaller than the point number, it is feasible to model long-range dependencies between regions with low computational cost by attention operations.
Several approaches have been explored for partitioning point clouds into regions, such as uniform grid partitioning \cite{lai2022stratified}, optimization-based methods \cite{papon2013voxel}, and over-segmentation networks \cite{hui2021superpoint}. 
However, uniform grid partitioning cannot separate points belonging to different objects.
Optimization-based methods and over-segmentation networks require additional optimization or training time, thus impeding the efficiency of region extraction.

We design a plug-and-play region-based feature enhancement (RFE) module to enable fast region extraction and discriminative point feature learning. The RFE module consists of a semantic-spatial region extraction (SSRE) part and a region dependency modeling (RDM) part.
In the SSRE part, we divide the input point cloud into a number of regions according to their semantic and spatial proximity.
We first assign points to different semantic groups according to their similarity in the feature space.
Nonetheless, semantic-level groups are too coarse to represent more complex scenes with shape details.
To address this, we cluster points within each semantic group into spatially distinct and non-overlapping regions according to their coordinates.
By taking advantage of the point features from deep segmentation models, we achieve more efficient and robust region extraction without an extra over-segmentation network or optimizer.
In the RDM part, we enhance region features by inter-region semantic and spatial correlations.
We first obtain region embeddings through average pooling over point features in each region and then perform region-level self-attention.
Thus the effective receptive field can be enlarged to the entire scene while maintaining a low computational cost.  
Subsequently, we fuse the point features and region features to obtain more accurate predictions, as shown in Fig.~\ref{fig:confusion} (c).

As a feature augmentation tool, our RFE module can be plugged into either voxel or point-based models and improve the segmentation results by capturing long-range context in large-scale point clouds. 
We conduct experiments on ScanNetV2 \cite{dai2017scannet} and S3DIS \cite{armeni20163d} datasets, and evaluate our RFE module on multiple 3D segmentation baselines.
Our REFL-Net achieves a $1.8\%$ mIoU gain on ScanNetV2 and $1.7\%$ mIoU gain on S3DIS, respectively.

In summary, this paper has the following contributions. 

\begin{itemize}
    \item We propose an intermediate region representation for efficient long-range context modeling for large-scale point clouds. Each region consists of a number of points close in both semantic and spatial space. Compared with points or voxels, the region representation is more compact, which is friendly for calculating pairwise correlations.
    
    \item We design a region-based feature enhancement module that is capable of modeling long-range context efficiently by clustering points into regions and exploiting their relationships. The proposed RFE module is plug-and-play and has strong generalization ability with various semantic segmentation models.
    
    \item We achieve state-of-the-art performance on point cloud semantic segmentation of complex scenes. Experiments on multiple baseline models and datasets show the superiority of our method, which achieves 1.8\% mIoU and 1.7\% mIoU gains on ScanNetV2 and S3DIS datasets.
\end{itemize}

\section{Related Work} \label{section:Related Work}

\subsection{Indoor Scene Semantic Segmentation}
\label{related:indoor_segmentation}

Previous indoor scene semantic segmentation methods can be categorized into voxel-based and point-based methods according to the data representation. Voxel-based methods \cite{graham2015sparse, graham2017submanifold, graham20183d, choy20194d} divide the input point cloud into regular voxels and process them using sparse convolutions. 
Though they have achieved great improvement, local shape details are inevitably ignored during voxelization.
Point-based methods \cite{pointnet, qi2017pointnet++, li2018pointcnn, thomas2019kpconv, wu2019pointconv, wang2019dynamic, hu2020randla, engelmann2020dilated, hu2020jsenet, lei2020seggcn, yan2020pointasnl, fan2021scf, xu2021paconv, zhao2021point, engel2021point, lai2022stratified, tang2022contrastive, ran2022surface, qian2022pointnext, ma2022rethinking} take the entire points as input to alleviate information loss during voxelization and learn point-wise features from their color and location attributes. 
Since aggregating information from local neighborhoods is extremely important for point feature learning, various feature aggregation kernels have been developed, mainly considering the weight assignment strategy.
PointNet++ \cite{qi2017pointnet++} and its variants\cite{qian2022pointnext, ma2022rethinking} use the max-pooling function to progressively aggregate point features, which ignores lots of local information.
To better capture local details,
some works like PointCNN \cite{li2018pointcnn}, KPConv \cite{thomas2019kpconv}, and PointConv \cite{wu2019pointconv} design various convolutional kernels that aggregate local points features according to the kernel weights, which are usually determined by the spatial distances between points.

Recently, inspired by Vision Transformer \cite{dosovitskiy2020image} and its wide applications in 2D images~\cite{yue2021vision, wang2021pyramid}, several approaches have been proposed to explore both semantic and spatial correlations between points by applying attention operations on point sets~\cite{zhao2021point, guo2021pct, engel2021point, chen2021HAPGN, lai2022stratified}.
However, these attention-based models typically limit the number of input points or neighborhood size due to the high computational cost that scales with the number of queries and keys, degrading their generalization ability to large-scale point clouds.
To enlarge the receptive field of attention layers under limited computation resources, Stratified Transformer \cite{lai2022stratified} proposes a stratified key-sampling strategy that samples nearby points densely and distant points sparsely as keys.
Nonetheless, the effective receptive field still cannot expand to the entire point cloud for large-scale indoor scene semantic segmentation.
Constrained by computation resources, voxel-based and point-based methods are restricted to aggregate features within a small local area, making it challenging to model distant contexts, such as the global layout and correlation between different objects.
To fully exploit these long-range contexts, we introduce regions as an intermediate representation to enable efficient region-level relationship modeling for large-scale point clouds, thereby enhancing point features for more accurate and robust predictions.

\subsection{Point Cloud Over-Segmentation}

Point cloud over-segmentation \cite{papon2013voxel, lin2018toward, landrieu2018large, landrieu2019point, hui2021superpoint} has been studied for many years to generate super-points or super-voxels as a more compact representation for point clouds and reduce the computational cost of downstream tasks.
Both unsupervised and supervised models have been proposed.
As an unsupervised method, the voxel cloud connectivity segmentation method (VCCS) \cite{papon2013voxel} generates super-points that conform to object boundaries by applying local $k$-means clustering in a hand-crafted feature space, which consists of the spatial coordinates, colors, and fast point feature histograms (FPFH) \cite{rusu2009fast}.  
However, VCCS is sensitive to the initial seed points.
Lin \emph{et al.} \cite{lin2018toward} formulate the over-segmentation task as a subset selection problem, which does not require seed point initialization. 
FPFH is also used as the point feature in this method \cite{lin2018toward}.
Nonetheless, hand-crafted features are not discriminative enough.
SSP \cite{landrieu2019point} is a supervised method that learns point embeddings with high contrast at object boundaries using a novel graph-structured contrastive loss.
However, SSP is not an end-to-end framework while using an optimization-based method \cite{guinard2017weakly} to generate super-points.
SPNet \cite{hui2021superpoint} is an end-to-end framework that learns association maps between points and super-point centers in the feature embedding space and updates the super-point centers and association maps iteratively. 
The final clustering results are obtained by performing the argmax function upon the final association maps. 
Different from the above methods, we introduce a simple yet effective clustering method that generates regions by utilizing the feature and spatial distance between points. By directly using the prediction results of the backbone semantic segmentation network, we can achieve fast region extraction without additional over-segmentation networks or optimization procedures.

\section{Our Method} \label{section:Method}

\begin{figure*}[!t]
  \centering  
  \vspace{-8mm}
  \includegraphics[width=1.0\linewidth]{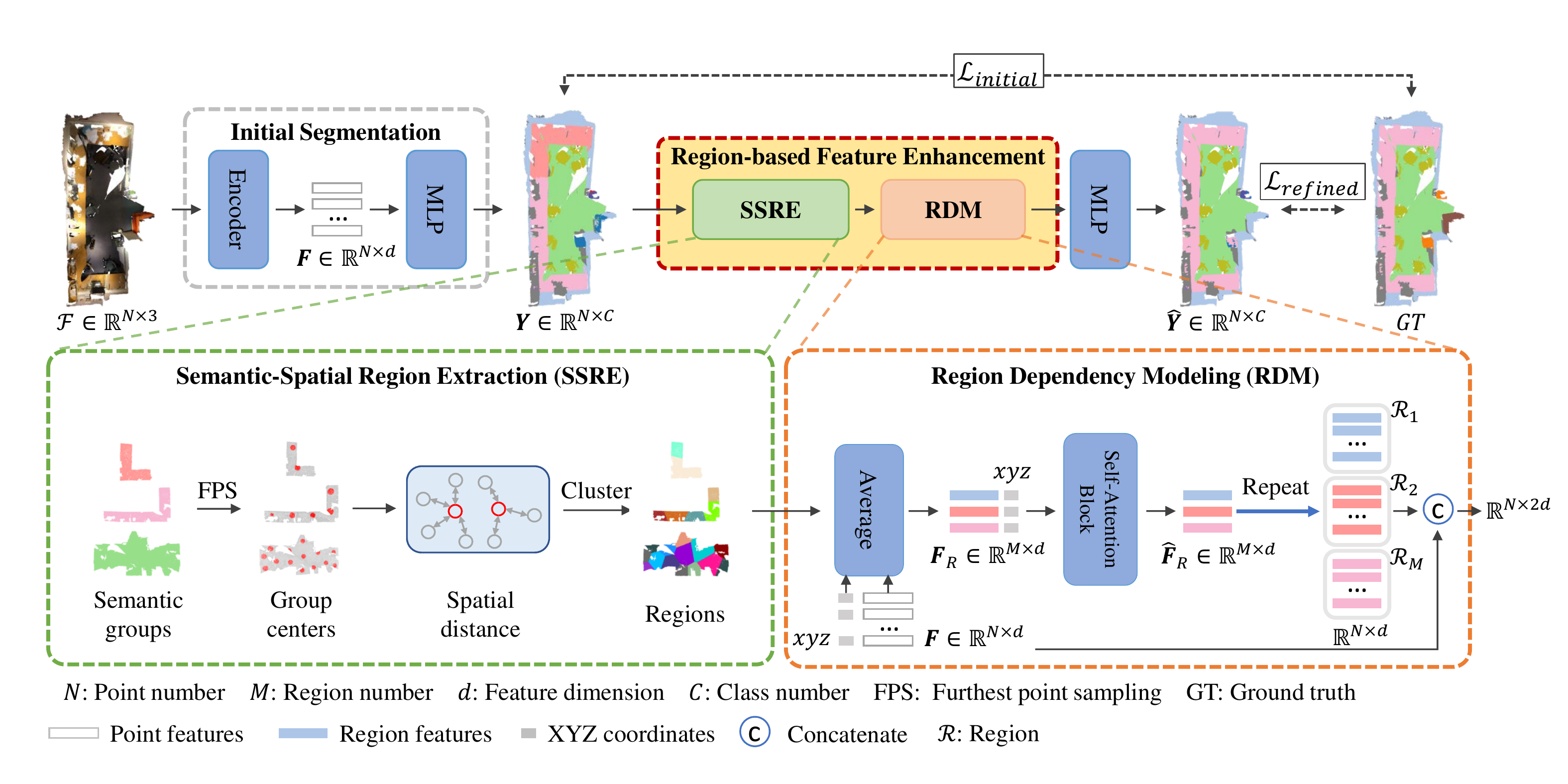}
  \vspace{-8mm}
  \caption{Overview of our REFL-Net for point cloud semantic segmentation. With a general segmentation backbone that extracts point features and makes the initial prediction, our RFE module extracts region-level context to enhance the point features for better semantic segmentation. 
  The RFE module consists of Semantic-Spatial Region Extraction (SSRE) and Region Dependency Modeling (RDM). The SSRE module takes the initial predictions as input and separates the point cloud into a set of local regions. The RDM module models the semantic and spatial correlations between regions and concatenates point features and region features for the final prediction.}
  \label{fig:our_pipeline}
\vspace{-4mm}
\end{figure*}

Our REFL-Net consists of a plug-and-play module named Region-based Feature Enhancement, which can be integrated with either voxel or point-based semantic segmentation models to explore long-range dependencies between different regions in indoor scenes.
The main architecture of our REFL-Net is shown in Fig.~\ref{fig:our_pipeline}.
Given a point cloud as input, we first utilize a semantic segmentation backbone that consists of an encoder to extract point features and a multi-layer perception to produce the initial segmentation.
Then we cluster points into a set of regions according to their semantic and spatial proximity in the semantic-spatial region extraction step. 
After that, we employ a self-attention block to model inter-region correlations in the region dependency modeling step.
Finally, we concatenate point features and corresponding region features to make per-point predictions.

\subsection{Network Architecture}
\label{sec:pipeline}

Given a point cloud $\mathcal{P}$ with $N$ points associated with RGB values $\mathcal{F}\in \mathbb{R}^{N\times 3}$ and $(x,y,z)$ coordinates $\mathcal{C}\in \mathbb{R}^{N\times 3}$, the segmentation task aims to predict the semantic category for each point $p_i\in \mathcal{P}$.
A basic semantic segmentation framework consists of an encoder that gradually extracts point features $\bm{F}\in \mathbb{R}^{N\times d}$, where $d$ is the feature dimension, and an MLP layer that predicts the probability vector $\bm{y}_i \in\mathbb{R}^{C} $ of $C$ object categories for point $p_i$. 

Within the basic semantic segmentation framework, the encoder aggregates information in a local neighborhood for each point separately. Though the encoder can learn multi-scale features by progressively increasing neighborhood size, it is non-trivial to distinguish points that have similar local geometry and appearance due to the lack of long-range context.
While representing long-range dependency in the entire scene with pairwise point correlations is extremely expensive in memory and computation, we address this confusion issue by using region-based feature enhancement. We first group the input points into a set of regions and then model long-range dependency at the region level, which is much more efficient and robust. 
Specifically, as a feature enhancement module, our RFE module is appended after the semantic segmentation network, taking the predicted probability vectors $\bm{Y}\in \mathbb{R}^{C}$ and point features $\bm{F}\in \mathbb{R}^{N\times d}$ as input.
The intermediate representation of the point cloud is composed of region features, noted as $\bm{F}_R\in \mathbb{R}^{M\times d}$, where $M$ is the number of extracted regions in the scene.
The output of our RFE module is generated by concatenating point features with their corresponding region features.
We design a semantic-spatial region extraction module to extract regions and a region dependency modeling module to learn region correlations.

\subsection{Semantic-Spatial Region Extraction} \label{sec:pcd}

Points in a local area usually have similar local features, which are greatly redundant and bring a huge cost in calculating pairwise point correlations.
Clustering points into more compact regions can effectively reduce the number of visual items and make distant relationship modeling more efficient.
In this part, we partition the input point cloud $\mathcal{P}$ into a set of regions $\{\mathcal{R}_1, \mathcal{R}_2, ...,\mathcal{R}_M\}$ while ensuring the semantic consistency and spatial proximity of points within each region.

\begin{figure}[!t]
  \centering
  \includegraphics[width=1.0\linewidth]{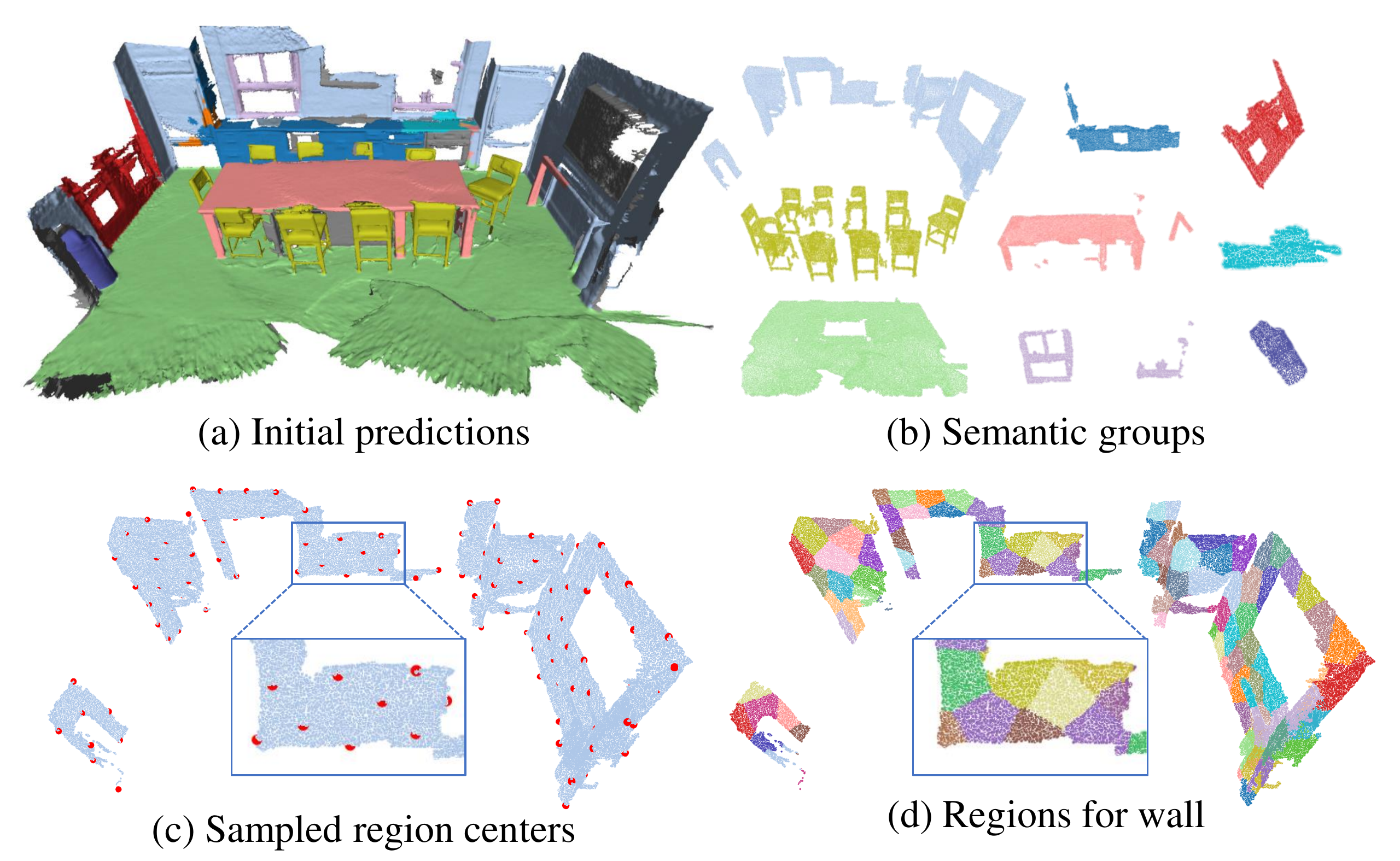}
  \caption{Semantic-spatial region extraction. From the initial semantic segmentation results (a), the point cloud is divided into a number of semantic groups (b). Then, based on the Euclidean distance between points and the uniformly sampled region centers (c), each group is split into fine-grained regions (d).}
\vspace{-4mm}
  \label{fig:ssre}
\end{figure}

To obtain high-quality over-segmentation results, the extracted regions are required to be semantically unitary, which means that regions cannot span objects belonging to multiple categories.
To preserve the semantic purity, we first divide the input points into a set of semantic groups according to the initial semantic predictions $\bm{Y}$.
Specifically, we assign each point $p_i$ to the category that has the maximum category probability. Then we gather all the points assigned to the same category into a group, resulting in $C$ point groups, as Fig.~\ref{fig:ssre} (b) shows.
In the semantic-only grouping step, a semantic group might contain points that are far apart in a large-scale scene. This coarse-grained grouping does not consider spatial continuity and is too coarse to model middle-level correlations in complex scenes.
Therefore, we further decompose each semantic group into a set of local regions considering the spatial continuity and region size. 
We adopt the furthest point sampling (FPS) \cite{qi2017pointnet++} method in each semantic group to sample a set of points as region centers, noted as $\{\hat{p}_1,\hat{p}_2,...,\hat{p}_{M_k}\}$, where $M_k$ is the number of regions in the $k$-th semantic group, as Fig.~\ref{fig:ssre} (c) shows.
To ensure consistent region sizes in various scenes, we use a hyper-parameter $s$ to control the maximum number of points in each region.
Then the number of regions is determined by $M_k = \lfloor N_k/s\rfloor$, where $N_k$ is the number of points in the $k$-th semantic group.
Finally, we assign each point $p_i$ to the group of its nearest center $\hat{p}_j$, according to their spatial distance $\lVert\bm{c}_i-\hat{\bm{c}}_j\rVert$.
The final semantic-spatial grouping result is a Voronoi tessellation of each semantic group in the 3D space, as shown in Fig.~\ref{fig:ssre} (d).

Through the semantic-spatial region extraction part, we can easily represent a point cloud as a set of regions, each representing a group of points that are semantically and spatially coherent.
Besides, the number of regions is much smaller than the number of points, which enables more efficient long-range context modeling.

\subsection{Region Dependency Modeling} \label{sec:rm}

After the semantic-spatial region extraction, our region dependency modeling module explores semantic and spatial correlations between different regions.
Unlike local geometric and appearance features, such global correlations incorporate both the relative positions within a scene and the co-occurrence of objects, which are beneficial for learning discriminative point features.

To begin with, we initialize region features $\bm{F}_R\in \mathbb{R}^{M\times d}$ and coordinates $\mathcal{C}_R\in \mathbb{R}^{M\times 3}$ by using the average of features and coordinates for points in each region.
After that, we treat each region as a query, all the other regions as keys, and compute region correlations using a self-attention module as 
\begin{equation}
\label{eq:attention}
    \begin{split}
    &\bm{Q}=\bm{F}_R \bm{W}_Q, \; \bm{K}=\bm{F}_R \bm{W}_K, \; \bm{V}=\bm{F}_R \bm{W}_V, \\
    &\bm{A} = \mathrm{Softmax}(\bm{QK}^T/\sqrt{d}), \\
    &\hat{\bm{F}}_R = \mathrm{MLP}(\bm{A}\bm{V}),
    \end{split}
\end{equation}
where $\bm{W}_Q, \bm{W}_K, \bm{W}_V\in \mathbb{R}^{d\times d}$ are learnable parameters, $\bm{A}\in \mathbb{R}^{M\times M}$ is the attention matrix, and $\hat{\bm{F}}_R\in \mathbb{R}^{M\times d}$ denotes the final enhanced region features.

To utilize the spatial context, we further apply relative positional encoding in the self-attention layers based on the $M$ region center coordinates.
Relative positional encoding has been well studied and can be divided into bias-mode relative positional encoding \cite{liu2021swin, wu2021rethinking, zhang2022pvt} and contextual relative positional encoding \cite{lai2022stratified}. 
We adopt the contextual relative positional encoding approach in Stratified Transformer\cite{lai2022stratified} for 3D point clouds to model the long-range dependency between regions because it can learn more adaptive positional features across various contexts.
Different from the relative positions between points, we compute the relative positions $\bm{r}_{i,j}=\hat{\bm{c}}_i - \hat{\bm{c}}_j$ between the centers of two regions to represent the distant spatial correlations across the whole scene. 
A learnable look-up table $\bm{T}\in \mathbb{R}^{3\times L\times d}$ is maintained to map the relative coordinates of two regions to the corresponding positional encoding.
$L$ is the number of quantized bins while we quantize the coordinates evenly with an interval $t$ for each dimension of the $x,y,z$ coordinates. 
The relative coordinates $\bm{r}_{i,j}$ are mapped into the indices $\bm{I}\in \mathbb{R}^{3\times M\times M}$ as
\begin{equation}
    \begin{split}
    &\bm{I}_{i,j}=\lfloor \bm{r}_{i,j}/t \rfloor, \\
    &\bm{pe}_{i,j}=\bm{T}_x[\bm{I}_{i,j}^x]+\bm{T}_y[\bm{I}_{i,j}^y]+\bm{T}_z[\bm{I}_{i,j}^z],
    \end{split}
\end{equation}
where $\bm{pe}_{i,j}\in \mathbb{R}^{d}$ denotes the relative positional embedding between region centers $\hat{p}_i$ and $\hat{p}_j$.
Finally, the positional bias is obtained from these positional embeddings and added to the attention map $\mathbf{A}$ to compute the region features $\hat{\bm{F}}_R$ in Eq.~(\ref{eq:attention}), same as \cite{lai2022stratified}.

We duplicate each region feature $\hat{\bm{F}}_{R,j}$ and concatenate it with point feature for points grouped in the region $\mathcal{R}_j$.
The concatenated point features $\hat{\bm{F}}\in \mathbb{R}^{N\times 2d}$ are fed into an MLP layer to make the final semantic predictions $\hat{\bm{Y}}\in \mathbb{R}^{N\times C}$, as shown in Fig.~\ref{fig:our_pipeline} (b).
Through the region dependency modeling part, the long-range context of the entire scene can be captured efficiently to help learn discriminative features.

\textbf{Discussion on Computational Complexity.} 
Using the intermediate region representation, we can reduce the computational complexity of the RDM part to $O(M^2)$, which is far less than the complexity $O(Nk)$ of window attention in StratifiedFormer\cite{lai2022stratified} since $M\ll N$, where $N$ is the number of input points and $k$ denotes the average number of points scattered in each window. Besides, different from StratifiedFormer which stacks 18 window attention layers, our RDM with only a three-layer self-attention block is lightweight, significantly speeding up the inference process. 

\subsection{Overall Objective}

Cross-entropy loss is utilized to optimize the entire semantic segmentation network. Our training procedure consists of two stages. First, a semantic segmentation model is pre-trained to generate the initial point cloud segmentation $\bm{Y}$. The loss function can be expressed as
\begin{equation}
\mathcal{L}_{initial}={\rm CrossEntropy}(\bm{Y}, \bm{G}),
\end{equation}
where $\bm{Y}$ represents the predicted class probability vector and $\bm{G}$ indicates the ground-truth class label of each point.
In the second stage, we jointly train the segmentation backbone and our region dependency modeling module. The cross-entropy loss is applied to the refined semantic predictions as
\begin{equation}
\mathcal{L}_{refined}={\rm CrossEntropy}(\hat{\bm{Y}}, \bm{G}),
\end{equation}
where $\hat{\bm{Y}}$ denotes the predicted class probability vector for all the points with region-enhanced features.

\section{Experiments} \label{section:Experiments}

We conduct extensive experiments to evaluate the efficiency and effectiveness of our REFL-Net by comparing several state-of-the-art approaches for indoor scene semantic segmentation. 
We also conduct ablation studies to demonstrate the long-range context modeling capability and strong generalization ability of our region-based feature enhancement module.

\subsection{Experimental Settings} \label{exp:setting}

\subsubsection{Data and Metric}  
We conduct experiments of point cloud semantic segmentation on two indoor scene datasets, ScanNetV2 \cite{dai2017scannet} and S3DIS \cite{armeni20163d}.
The ScanNetV2 dataset contains 1,201 training scenes, 312 validation scenes, and 100 test scenes. There are 20 semantic categories in total. The S3DIS dataset consists of scanned point clouds of 271 rooms in six areas from three buildings. According to the general protocol \cite{pointnet}, Area 5 is used as the test set, and other areas are used as the training set, with 13 semantic categories in total. 
We use mean class-wise intersection over union (mIoU) as the evaluation metric for the semantic segmentation results.
All experiments were performed on a single Tesla V100 GPU.

\subsubsection{Implementation Details}

Based on the overall architecture shown in Fig.~\ref{fig:our_pipeline}, we have various options for the encoder, region extraction module, and region dependency modeling module. For the initial segmentation before the semantic-spatial region extraction step, we pre-train a semantic segmentation model to produce the initial predictions. 
Specifically, we train MinkowskiUNet34C~\cite{choy20194d} for 120k iterations by utilizing the SGD optimizer with learning rate  as $1e^{-1}$ and weight decay as $1e^{-4}$ respectively. The batch size is 10. Following StratifiedFormer\cite{lai2022stratified}, we use RGB values as inputs and apply the same data augmentation including random scaling, random rotation, drop color, chromatic jitter, etc. 
Besides, we voxelize the input point cloud with a voxel size of $0.02m$ to reduce the computational and storage burden for large-scale scenes, following previous work~\cite{choy20194d,lai2022stratified}.

In the region extraction step, we set the region size $s$ to 200 points per region, leading to nearly $100$ regions per scene. 
In the region dependency modeling part, we adopt a three-layer attention block, with the feature dimension $d=128$ and head number as $8$.
For the contextual relative position encoding, we initialize the look-up table for relative positional embedding according to \cite{lai2022stratified}. We set the maximum relative distance to $2m$ and the quantization interval $t=0.02m$.
In the joint training stage, we use the SGD optimizer with learning rate as $5e^{-3}$ and $5e^{-4}$ for the segmentation model and region dependency modeling module respectively. The batch size is set to 4. 
Note that different scenes may contain different numbers of regions $M$. Therefore, the attention matrix $\bm{A}\in \mathbb{R}^{M\times M}$ has different dimensions for different scenes. We process the region features and compute the attention matrices for each scene in a batch sequentially in each iteration.
We also implement a learning rate warm-up process with 3,000 iterations for a more stable training process.

\begin{table*}[!t]
\caption{Semantic segmentation mIoU on the ScanNetV2 and S3DIS datasets, model parameters, and inference performance. The value with the best result is highlighted in \textbf{bold}. The value with the second-best result is highlighted with an \underline{underline}.}\label{tab:main_result}
\centering
\begin{threeparttable}
\begin{tabular}{l|c|c|c|c|c|c}
    \hline
    Methods&ScanNetV2 (\%) & S3DIS Area5 (\%) & S3DIS 6-Fold (\%)   & Time (s) & Params (M) & FLOPs (G)\\
    \hline
    PointNet\cite{pointnet}& - & 41.1 & 47.6  & - & 3.6 & 35.5 \\
    PointNet++\cite{qi2017pointnet++} & 53.5 &53.5 & 54.5  & - & 1.0 & 7.2 \\
    PointCNN\cite{li2018pointcnn}& - & 57.3 & 65.4  & - & 0.6 & - \\
    KPConv\cite{thomas2019kpconv}& 69.2 & 67.1 & 70.6  & 118 & 15.0 & - \\
    PointNeXt-XL\cite{qian2022pointnext} & 71.5 &70.5 & \underline{74.9}  & 1,050 & 41.6 & 84.8 \\
    StratifiedFormer\cite{lai2022stratified} &\textbf{74.3} &\textbf{72.0} & -   &1,821 &18.8 & - \\
    \hline
    MinkowskiNet\cite{choy20194d}&72.2 &65.8 &-  & 82 & 37.9 & - \\
    REFL-Net-Mink (Ours) & \underline{74.0} & 66.9 &-  &217 &40.2 &- \\
    \hline
    PointTransformer \cite{zhao2021point} &70.6&70.4 & 73.5  & 784 &7.8 &5.6 \\
    REFL-Net-PT (Ours) &71.5 &\underline{71.4} &\textbf{75.2}  & 1,055 &8.8 &9.0 \\
    \hline
\end{tabular} 
\end{threeparttable}
\vspace{-4mm}
\end{table*}

\subsection{Comparison with State-of-the-art Methods} \label{exp:comparison}

We compare our method with previous works~\cite{pointnet, qi2017pointnet++,li2018pointcnn,thomas2019kpconv} and state-of-the-art methods, PointNeXt~\cite{qian2022pointnext}, StratifiedFormer~\cite{lai2022stratified}, MinkowskiNet~\cite{choy20194d} and PointTransformer~\cite{zhao2021point}. The semantic segmentation results for the ScanNetV2 and S3DIS datasets are presented in Table \ref{tab:main_result}.
In addition to mIoUs, we also report the inference times for a group of scenes in the ScanNetV2 dataset that contains an average of $10^8$ points on a Tesla V100 to evaluate the efficiency of different models.
To further compare model complexity, we report the number of parameters and the floating point operations (FLOPs) of these models in Table \ref{tab:main_result}.

We implement two variants of our REFL-Net, i.e., REFL-Net-Mink and REFL-Net-PT, using two semantic segmentation backbones (MinkowskiNet \cite{choy20194d} and PointTransformer~\cite{zhao2021point}). 
On the ScanNetV2 validation set, our REFL-Net-Mink obtains comparable mIoU (74.0\%) with the state-of-the-art result (74.3\%) of StratifiedFormer \cite{lai2022stratified}, with $9\times$ speedup. 
Compared with StratifiedFormer which directly uses window-based attention on points, our REFL-Net performs self-attention on regions, which is much more efficient.
Our REFL-Net-PT, which integrates the proposed RFE module with PointTransformer~\cite{zhao2021point}, obtains 1.0\% mIoU increase on the S3DIS Area $5$ and 1.7\% mIoU increase on the S3DIS 6-Fold cross-validation experiment, with only a slight increase in model size ($7.8M\to 8.8M$) and FLOPs ($5.6G\to 9.0G$).
These results demonstrate that our REFL-Net can effectively improve semantic segmentation performance with a small increment in model complexity.
 
\begin{figure*}[!t]
  \centering
  \includegraphics[width=1.0\linewidth]{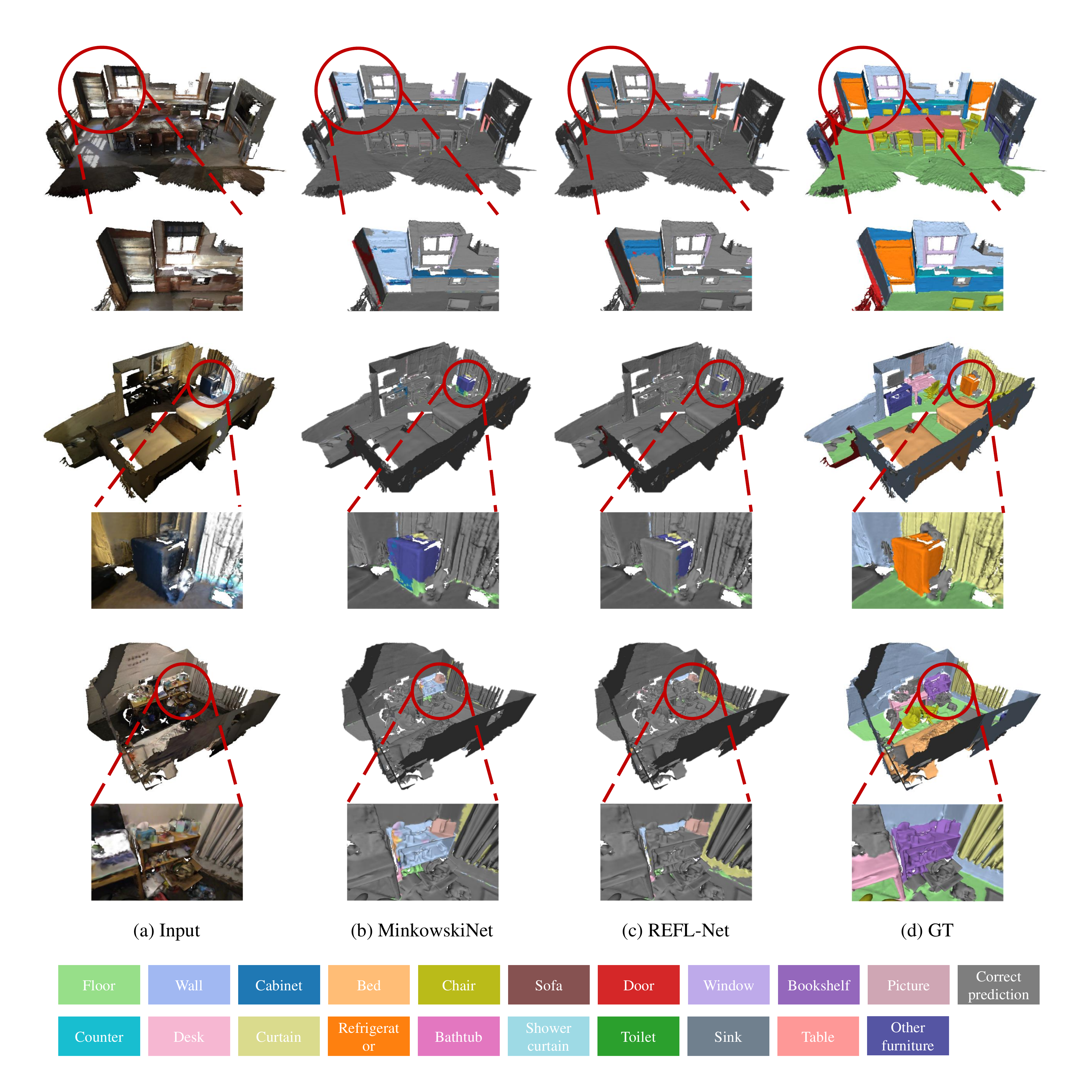}
  \vspace{-4mm}
  \caption{Semantic segmentation results of three indoor scenes in ScanNetV2. (a) The input point clouds. We visualize the semantic segmentation errors (colored points) of MinkowskiNet \cite{choy20194d} and our REFL-Net in (b) and (c), respectively. Grey points are correctly classified compared with the ground truth (d).}
  \label{fig:Visualization}
\end{figure*}

The performance gains of the two variants of our REFL-Net compared with the backbone segmentation networks benefit from the proposed region-based feature enhancement module, which can effectively capture long-range dependencies in a scene and learn more distinctive point features.
We visualize the semantic segmentation error maps in Fig.~\ref{fig:Visualization}. 
As shown in Fig.~\ref{fig:Visualization} (b), the refrigerators in the first two scenes and the bookshelf in the third scene are incorrectly classified using MinkowskiNet due to the indistinguishable local appearances.
Utilizing the proposed RFE module, our REFL-Net significantly reduces the prediction errors by perceiving larger-scale surrounding environments, as shown in Fig.~\ref{fig:Visualization} (c).
 
To further analyze the segmentation results across different categories, we present the mIoU of our REFL-Net-Mink and MinkowskiNet~\cite{choy20194d} on the ScanNetV2 test set in Table \ref{tab:iou_per_class}. We train the two models using the ScanNetV2 training set. Following MinkowskiNet, we employ rotation averaging as the augmentation strategy during the evaluation phase.  
Compared to MinkowskiNet, our REFL-Net-Mink achieves a 1.5\% increase in mIoU and generally demonstrates improved performance across various categories. With the long-range context learned by the RFE module, our REFL-Net-Mink can better distinguish objects that are easily misclassified. For example, our REFL-Net-Mink demonstrates significant IoU improvement in categories such as `Refrigerator' (+14.9\%), `Curtain' (+7.3\%), `Other' (+6.5\%), `Shower curtain' (+5.7\%).

\begin{table}[!t]
\caption{Performance improvements by combining our region-based feature enhancement with different backbone models. The value with the best result is highlighted in \textbf{bold}.}
\label{tab:diff_baselines}
\centering
\begin{threeparttable}
\begin{tabular}{l|c|c}
    \hline
    Methods & Dataset & mIoU (\%)\\
    \hline
    MinkowskiUNet34C\cite{choy20194d} & ScanNetV2 & 72.2 \\
    + Our RFE & ScanNetV2 & \textbf{73.4} \\
    \hline
    MinkowskiUNet18\cite{choy20194d} & S3DIS & 65.8 \\
    + Our RFE & S3DIS & \textbf{66.5} \\
    \hline
    KPConv (deform)\cite{thomas2019kpconv} & SemanticKITTI & 62.9 \\
    + Our RFE & SemanticKITTI & \textbf{63.5} \\
    \hline
    KPConv (deform)\cite{thomas2019kpconv} & S3DIS & 67.3 \\
    + Our RFE & S3DIS & \textbf{67.7} \\
    \hline
    PointTransformer\cite{zhao2021point} & S3DIS & 70.4 \\
    + Our RFE & S3DIS & \textbf{71.4} \\
    \hline
    Point-NeXt-XL\cite{qian2022pointnext} & S3DIS & 70.5 \\
    + Our RFE & S3DIS & \textbf{71.6} \\
    \hline
\end{tabular}
\end{threeparttable}
\vspace{-4mm}
\end{table}

\begin{table*}[!t]
\caption{Per-class IoU(\%) and mIoU(\%) on the ScanNetV2 test set. Models for evaluation are trained using the ScanNetV2 training set. The value with the best result is highlighted in \textbf{bold}. \dag ~indicates results are reproduced by us.}\label{tab:iou_per_class}
\centering
\begin{threeparttable}
    \begin{tabular}{c|c|c|c|c|c|c|c|c|c|c}
    \hline
         & \textbf{mIoU} & Bathtub & Bed & Bookshelf & Cabinet & Chair & Counter & Curtain & Desk & Door  \\ \hline
        MinkowskiNet\dag~\cite{choy20194d} & 71.36  & \textbf{92.70}  & 78.20  & \textbf{81.50}  & 69.80  & 82.70  & \textbf{49.20}  & 79.20  & 66.40  & 61.30   \\ \hline
        REFL-Net-Mink & \textbf{72.87}  & 73.70  & \textbf{82.30}  & 76.60  & \textbf{72.60}  & \textbf{85.20}  & 46.80  & \textbf{86.50}  & \textbf{68.40}  & \textbf{63.40}   \\ \hline
        \hline
        Floor & Other & Picture & Refrigerator & Shower curtain & Sink & Sofa & Table & Toilet & Wall & Window  \\ \hline
        95.10  & 50.00  & \textbf{31.80}  & 62.40  & 71.70  & 72.40  & \textbf{77.80}  & 62.80  & \textbf{91.50}  & 84.40  & 66.20   \\ \hline
        \textbf{95.30}  & \textbf{56.50}  & 29.70  & \textbf{77.30}  & \textbf{77.40}  & \textbf{77.70}  & 74.90  & \textbf{66.60}  & 90.70  & \textbf{85.00}  & \textbf{70.80}   \\ \hline
    \end{tabular}
\end{threeparttable}
\end{table*}

\subsection{Generalization Ability of RFE Module}
\label{exp:generalization}

Our region-based feature enhancement module can be plugged into most semantic segmentation models.
To demonstrate its generalizability, we conduct experiments by appending the RFE module after different semantic segmentation backbones and show the results in Table \ref{tab:diff_baselines}.

Typically, point-based methods \cite{thomas2019kpconv, zhao2021point, qian2022pointnext} work better than voxel-based methods \cite{choy20194d}. 
For point-based methods, we select KPConv\cite{thomas2019kpconv}, PointTransformer \cite{zhao2021point} and Point-NeXt-XL\cite{qian2022pointnext} as the backbone models and conduct semantic segmentation on the S3DIS dataset and the SemanticKITTI dataset~\cite{behley2019semantickitti}.
For a fair comparison, we adopt the same data augmentation as the corresponding backbone models.
As shown in the last three rows of Table \ref{tab:diff_baselines}, our RFE module brings $0.4\%$, $1.0\%$ and $1.1\%$ mIoU gains on S3DIS Area 5 for KPConv\cite{thomas2019kpconv}, PointTransformer \cite{zhao2021point}, and Point-NeXt-XL\cite{qian2022pointnext}, respectively.
To demonstrate the effectiveness of our method on outdoor scenes, we also conduct experiments on the SemanticKITTI~\cite{behley2019semantickitti} dataset. The results are shown in the third row of Table \ref{tab:diff_baselines}. By augmenting the KPConv backbone with our RFE module, we achieve a 0.6\% mIoU increase on the validation set.
These methods consider point-wise correlations in local neighborhoods, making it hard to effectively capture long-range context.
In comparison, our region-based representation and dependency modeling method brings the benefit of perceiving locations and surrounding environments and results in more distinctive features.

For voxel-based methods, we choose MinkowskiNet~\cite{choy20194d} as the backbone for its good performance and efficiency. We conduct semantic segmentation on both ScanNetV2 and S3DIS datasets. The results are shown in the first two rows of Table \ref{tab:diff_baselines}. 
1.2\% and 0.7\% mIoU gains are respectively achieved for the two datasets, showing the effectiveness of our RFE module in capturing long-range context in complex scenes.

\subsection{Ablation Study} \label{exp:ablation}

\subsubsection{Analysis of Components in the RFE Module}
 
To demonstrate the importance of the semantic-spatial region extraction part and region dependency modeling part in our RFE module, we conduct a set of experiments on the ScanNetV2 dataset and report the results in Table \ref{tab:effect_of_components}.
In Table \ref{tab:effect_of_components}, SeRE stands for semantic region extraction, which clusters points with the same prediction into groups based on the initial segmentation results. 
SpRE represents spatial region extraction, which clusters points into different groups based on their spatial coordinates.
RPE indicates the relative positional encoding.

\begin{table}[!t]
\caption{Effect of components in our RFE module. The value with the best result is highlighted in \textbf{bold}.} \label{tab:effect_of_components} 
\centering
\begin{threeparttable}
\begin{tabular}{c|c|c|c|c|c}
    \hline
    \multirow{2}{*}{EXP} & RFE & RFE & RDM  & RDM  & \multirow{2}{*}{mIoU (\%)}\\
    &w SeRE&w SpRE& w/o RPE & w RPE &\\
    \hline
    \uppercase\expandafter{\romannumeral1} & - & - & - & - & 72.22 \\
    \hline
    \uppercase\expandafter{\romannumeral2} & \checkmark & - & - & - & 72.37\\
    \uppercase\expandafter{\romannumeral3} & - & \checkmark & - & - & 72.37\\
    \uppercase\expandafter{\romannumeral4} & \checkmark &\checkmark & - & - & 72.80\\
  \hline
      \uppercase\expandafter{\romannumeral5} & \checkmark & \checkmark & \checkmark & - & 73.07\\
    \uppercase\expandafter{\romannumeral6} & \checkmark & \checkmark & - & \checkmark & \textbf{73.41}\\
    \hline
\end{tabular}
\end{threeparttable}
\vspace{-4mm}
\end{table}

First, to analyze the effects of different components of the SSRE part, we adopt different region extraction strategies after the semantic segmentation model in EXP \uppercase\expandafter{\romannumeral2}, \uppercase\expandafter{\romannumeral3} and \uppercase\expandafter{\romannumeral4}. 
In EXP \uppercase\expandafter{\romannumeral1}, MinkowskiNet is adopted for semantic segmentation and serves as the baseline.
In EXP \uppercase\expandafter{\romannumeral2}, only semantic region extraction (SeRE) is used, resulting in 0.15\% mIoU gain.
This is because the regional features represent the semantic centers of different categories, which help pull the enhanced point features away from the classification plane, making the point classification easier.
In EXP \uppercase\expandafter{\romannumeral3}, spatial region extraction (SpRE) is adopted, also leading to 0.15\% mIoU gain.
This improvement can be attributed to the local consistency constraint, which forces points in the same local region to have similar features and makes predictions smoother.
In EXP \uppercase\expandafter{\romannumeral4}, we integrate the semantic and spatial region extraction and achieve 0.58\% mIoU gain.
While the semantic grouping is too coarse and may assign points to the wrong class and the spatial grouping cannot separate adjacent points belonging to different categories without semantic information, the improvement by semantic region extraction or spatial region extraction alone is limited. 
By combining them, we can extract regions that are more spatially and semantically coherent. 
 
To analyze the effect of region dependency modeling, we employ the RDM part after region extraction and show the results in EXP \uppercase\expandafter{\romannumeral5} and \uppercase\expandafter{\romannumeral6}.
We first evaluate the effect of our RDM module without relative positional encoding (RPE) in EXP \uppercase\expandafter{\romannumeral5}.
0.27\% mIoU gain is obtained compared with EXP \uppercase\expandafter{\romannumeral4}, demonstrating the effectiveness of inter-region correlation modeling.
Furthermore, we evaluate the effect of relative positional encoding in EXP \uppercase\expandafter{\romannumeral6}.
By applying relative positional encoding in the RDM part, we obtain an additional 0.34\% mIoU gain compared with EXP \uppercase\expandafter{\romannumeral5}, demonstrating the advantage of modeling long-range spatial relationships.

\begin{figure}[!t]
  \centering
  \includegraphics[width=1.0\linewidth]{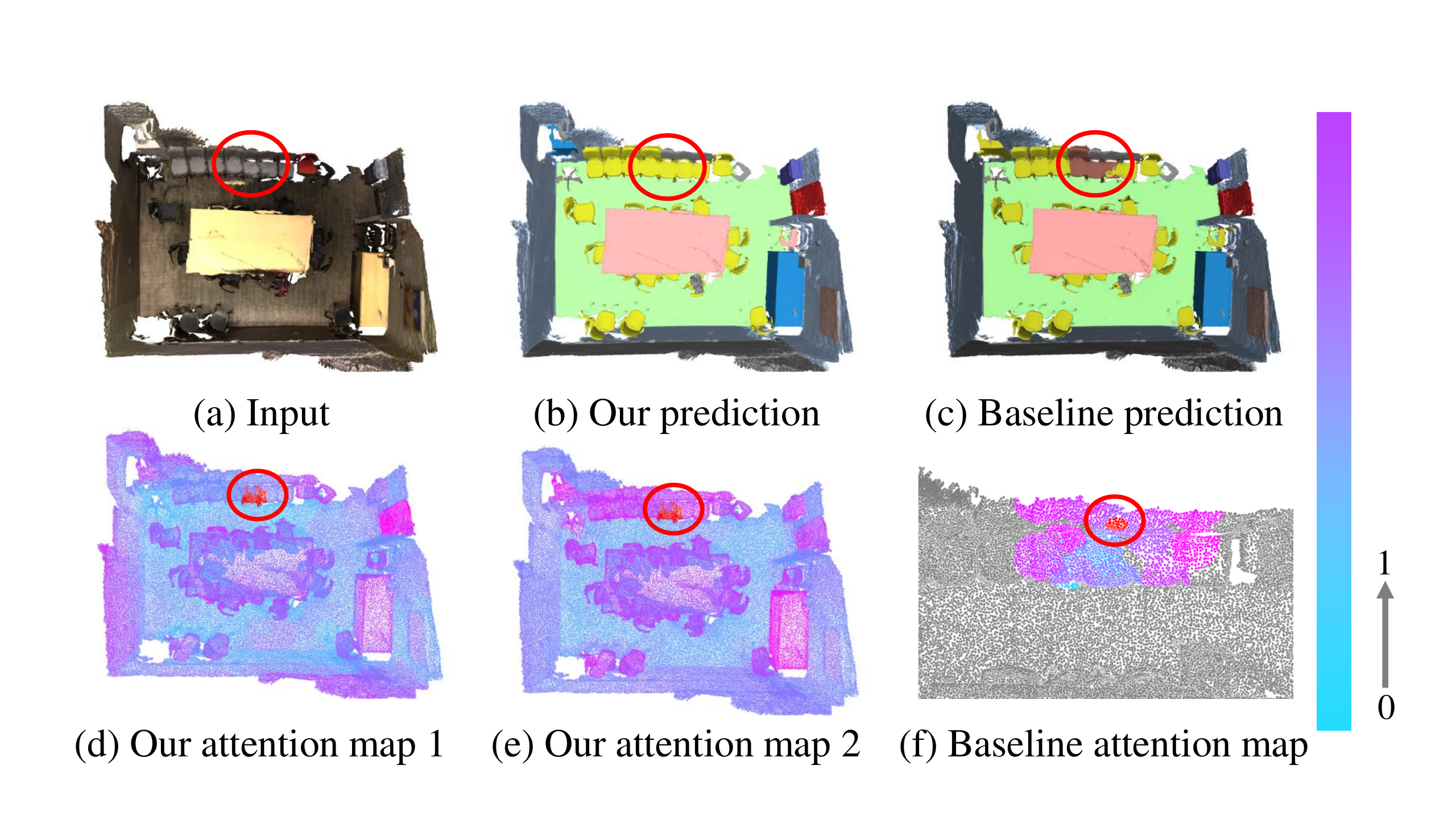}
  \vspace{-8mm}
  \caption{Attention map visualization. The query region is marked by the red circles. The two attention maps in our RDM module demonstrate the context of the scene layout (d) and surrounding objects (e). The point-wise attention in PointTransformer\cite{zhao2021point} only captures contextual information in a local neighborhood (f), leading to incorrect predictions for the middle chairs.}
  \label{fig:AttentionMap}
  \vspace{-4mm}
\end{figure}

\subsubsection{Visualization of Attention Maps}  
To verify the long-range context modeling ability of the RDM part, we visualize the attention maps of the self-attention layers that represent inter-region correlations.
Given the query region noted by a red circle, we select two representative attention maps to show the perception of scene layout and surrounding objects in Fig.~\ref{fig:AttentionMap} (d) and (e) respectively.  
In Fig.~\ref{fig:AttentionMap} (d), the `chair' region has higher attention weights to the `wall' representing the horizontal layout, which helps to perceive its global position in the scene.
In Fig.~\ref{fig:AttentionMap} (e), the `chair' region has higher attention weights to objects including tables and chairs, revealing co-occurring dependencies between multiple furniture objects in indoor scenes.  
The steadiness of correlations between object pairs such as `chair-wall' and `chair-table' in different scenes provides strong support for point classification when the local feature is indistinguishable.
We also visualize the prediction result and attention map of PointTransformer\cite{zhao2021point} in Fig.~\ref{fig:AttentionMap} (c) and (f).
While only performing point-wise attention in a local neighborhood, PointTransformer fails to predict the correct category for the chairs in the middle. 
In comparison, our RDM can effectively model the long-range context of the whole scene and learn more discriminative region features, resulting in more accurate predictions.
 
\subsubsection{Point Cloud Over-Segmentation}

\begin{figure}[!t]
  \centering
  \includegraphics[width=0.9\linewidth]{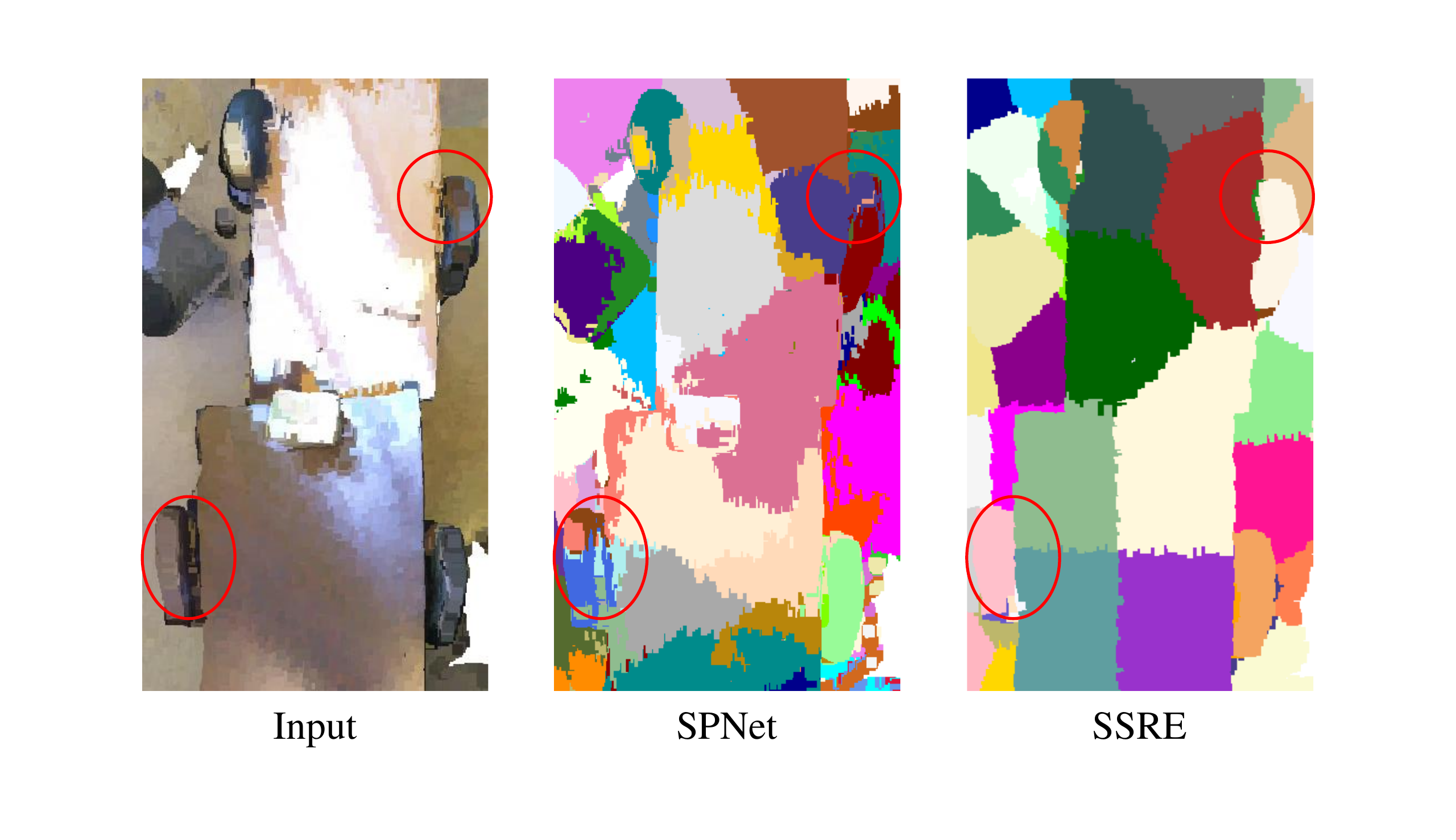}
  \vspace{-6mm}
  \caption{Region extraction results using SPNet \cite{hui2021superpoint} and our semantic-spatial region extraction (SSRE).}
  \label{fig:cluster_vis}
  \vspace{-4mm}
\end{figure}

We also compare different options for the point cloud over-segmentation in Table \ref{tab:comp_cluster}, including our semantic-spatial region extraction (SSRE) and the state-of-the-art SPNet \cite{hui2021superpoint}.
Since the number of generated regions cannot be deterministically specified for either of the two region extraction methods, we set the super-point ratio to 0.002 for SPNet and the region size to 200 for SSRE to ensure a comparable number of regions in each scene.
To evaluate the region extraction results, we calculate the average region entropy and boundary recall (BR) \cite{hui2021superpoint} on the ScanNetV2 validation set.
The region entropy reflects the semantic purity of each region and is defined as, 
\begin{equation}
    E(\mathcal{R}_j) = -\sum_{k=1}^{C}h_k log(h_k),
\end{equation}
where $h_k$ denotes the ratio of points belonging to the $k$-th class in each region $\mathcal{R}_j$ and $C$ is the total number of semantic classes.  
To measure the semantic purity of region extraction results on the entire validation set,
we compute the average region entropy across different scenes and regions as
\begin{equation}
    E_{val} = \frac{1}{N_S}\sum_{i=1}^{N_S}\big(\frac{1}{M_i}\sum_{j=1}^{M_i}E(\mathcal{R}_{ij})\big),
\end{equation}
where $N_S$ denotes the number of scenes in the validation set and $M_i$ is the number of regions in the $i$-th scene.
Boundary recall is computed in each scene and averaged over the validation set.
As Table \ref{tab:comp_cluster} shows, we achieve higher boundary recall and lower region entropy compared with SPNet.
In comparison with SPNet, our SSRE can be easily integrated into various segmentation models and requires no additional training on over-segmentation tasks.
We can benefit from a powerful feature extractor in the segmentation backbone and obtain better region extraction results, as shown in Fig.~\ref{fig:cluster_vis}.
Besides, we obtain 0.88\% and 1.19\% mIoU gains using our RFE module with SPNet and SSRE, respectively, compared with the 72.22\% mIoU of MinkowskiNet (the first row in Table~\ref{tab:effect_of_components}).
It shows the robustness of our region-based feature enhancement framework to different region extraction methods.
In addition, we evaluate our SSRE module with different spatial grouping algorithms, i.e., the furthest point sampling (FPS) and $k$-means clustering.
As illustrated in Table~\ref{tab:comp_cluster}, the comparable performance demonstrates the robustness of our method with different spatial region extraction methods.

\begin{table}[!t]
\caption{Comparison of different region extraction methods on ScanNetV2 validation set. The value with the best result is highlighted in \textbf{bold}.}
\label{tab:comp_cluster}
\centering
\begin{threeparttable}
\begin{tabular}{l|ccc|c}
    \hline
    Method & \# Region & Entropy & BR (\%) & mIoU (\%)\\
    \hline
    SPNet\cite{hui2021superpoint} & 318 & 0.185 & 67.78 & 73.10 \\
    SSRE-$k$means & 278 & 0.116  & 68.23 & 73.33 \\
    SSRE-FPS & 278 & 0.112 & 68.80 & \textbf{73.41} \\
    \hline
\end{tabular}
\end{threeparttable}
\vspace{-4mm}
\end{table}

\subsubsection{Analysis of Region Size} In the semantic-spatial region extraction part, the region size $s$ serves as a hyper-parameter and controls the maximum number of points in final regions.
We test different region sizes and the results on the ScanNetV2 validation set are shown in Table \ref{tab:ablation_region_size}.
By setting region size to 100 and 200, we get 0.56\% and 1.19\% mIoU gain respectively, compared with the baseline model.
A smaller region size results in more regions and correspondingly more complex and harder-to-capture correlations, thus bringing less gain than a larger region size.
As the region size increases to 300 and 400, improvements become smaller (0.67\% and 0.60\%), mainly because larger regions sacrifice more details and limit the segmentation performance.
In our final model, we set the region size $s$ to 200 for the best result.

\begin{table}[!t]
\caption{Analysis of region size on ScanNetV2 validation set. The value with the best result is highlighted in \textbf{bold}.}
\label{tab:ablation_region_size}
\centering
\begin{threeparttable}
\begin{tabular}{c|ccccc}
    \hline
    Region size & - & 100 & 200 & 300 & 400\\
    \hline
    mIoU (\%) & 72.22 & 72.78 & \textbf{73.41} & 72.89 & 72.82 \\
    \hline
\end{tabular}
\end{threeparttable}
\vspace{-4mm}
\end{table}

\subsubsection{Analysis of the Attention Layer Number}
In the region dependency modeling part, we utilize self-attention layers to explore inter-region correlations.
In order to determine a proper number of self-attention layers, we conduct several experiments on the ScanNetV2 validation set and report the results in Table \ref{tab:ablation_attn_layer}.
As the layer number increases from 1 to 3, the mIoU gets higher from 72.95\% to 73.41\% due to the increasing relation modeling ability.
However, too many self-attention layers may lead to over-fitting and limit the relation modeling capability, resulting in 73.07\% and 73.00\% mIoU respectively.
We also show the parameter number and the average inference time needed for processing $1,000$ region sequences, each with a length of 100, for the self-attention block.
Results in the second row of Table \ref{tab:ablation_attn_layer} show that our self-attention block is lightweight and fast.
In our final model, we use a three-layer self-attention block for the best result.

\begin{table}[!t]
\caption{Analysis of the attention layer number on ScanNetV2 validation set. The value with the best result is highlighted in \textbf{bold}.}
\label{tab:ablation_attn_layer}
\centering
\begin{threeparttable}
\begin{tabular}{c|ccccc}
    \hline
    \# Attention Layer & 1 & 2 & 3 & 4 & 5\\
    \hline
    Layer Params (M) & 0.6 & 1.2 & 1.8 & 2.4 & 3.0 \\
    Time (s) & 1.7 & 3.7 & 5.0 & 6.7 & 8.5 \\
    \hline
    mIoU (\%) & 72.95 & 73.28 & \textbf{73.41} & 73.07 & 73.00 \\
    \hline
\end{tabular}
\end{threeparttable}
\vspace{-4mm}
\end{table}

\subsection{Discussion} \label{sec:discussion}
Though our region-based feature enhancement module significantly reduces prediction errors by enhancing point features with long-range contexts, it sometimes fails to segment ambiguous points, especially at region boundaries of different object categories. For example, in the first row of Fig.~\ref{fig:Visualization} (c), the top of the refrigerator is categorized as `cabinet' by mistake, and the right portion of the refrigerator in the second row is misclassified as the `other' category. As we combine region features with point features from the backbone model to produce the final prediction, the resulting segmentation may still contain inaccuracies if the initial point features are significantly misleading.

Another limitation of our RFE module is its uniform partitioning of each semantic group based on point numbers. This may cause different region granularity when the number of points differs significantly across different categories, such as `wall' and `chair'. Consequently, regions of the categories with fewer points may ignore shape details, resulting in less representative region features. A possible solution is to partition regions adaptively based on local geometric diversities, such as planes and edges. Investigating adaptive region divisions may be a worthwhile direction for future research.
 
\section{Conclusion} \label{section:Conclusion}
In this paper, we introduce a point feature enhancement framework, REFL-Net, using the region representation for point cloud segmentation in large-scale scenes. 
As a key component of our REFL-Net, the proposed region-based feature enhancement module extracts local regions as the intermediate representation and models long-range region dependencies to augment point features. 
With the proposed region representation, our RFE module effectively models long-range context in large-scale indoor scenes and achieves significant improvements with much less memory and computation cost than point-wise correlation frameworks. 
As a plug-and-play module, our RFE module can be flexibly integrated with various semantic segmentation backbones, including both point-based and voxel-based networks.
A future direction is to extend our region-based feature enhancement framework to semantic segmentation of high-resolution images or high-volume videos to capture object correlations more efficiently. 

\bibliographystyle{IEEEtran}
\bibliography{sample-based.bib}

\vspace{-50mm}
\begin{IEEEbiography}
[{\includegraphics[width=1in,clip,keepaspectratio]{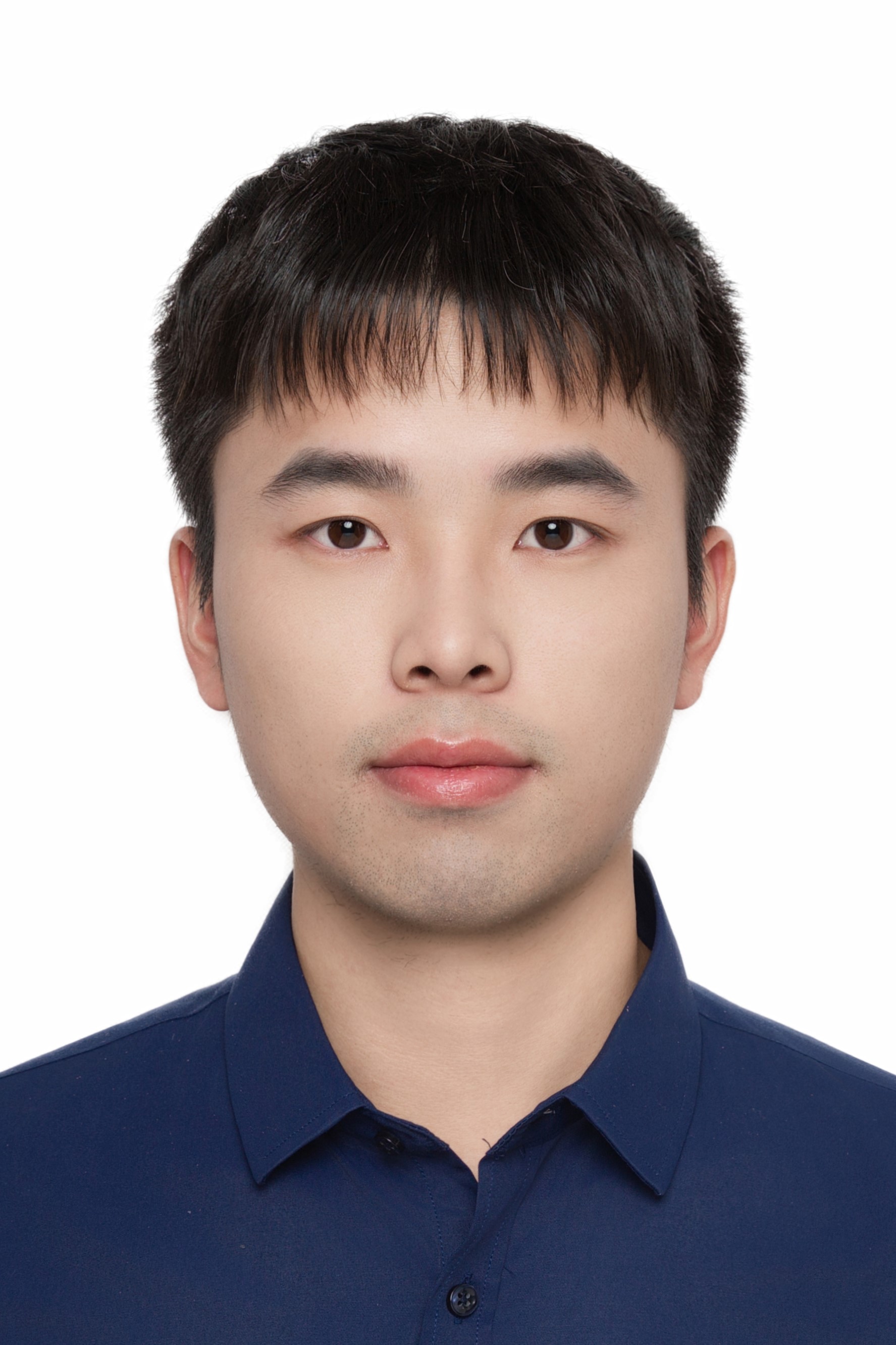}}]
{Xin Kang} received the B.S. degree in department of electronic engineering and information science from University of Science and Technology of China, in 2020. He is currently studying for Ph.D. being advised
by Xuejin Chen in USTC. 
His research interest involves 3D computer vision, especially scene semantic segmentation and scene generation.
\end{IEEEbiography}
\vspace{-50mm}

\begin{IEEEbiography}
[{\includegraphics[width=1in,clip,keepaspectratio]{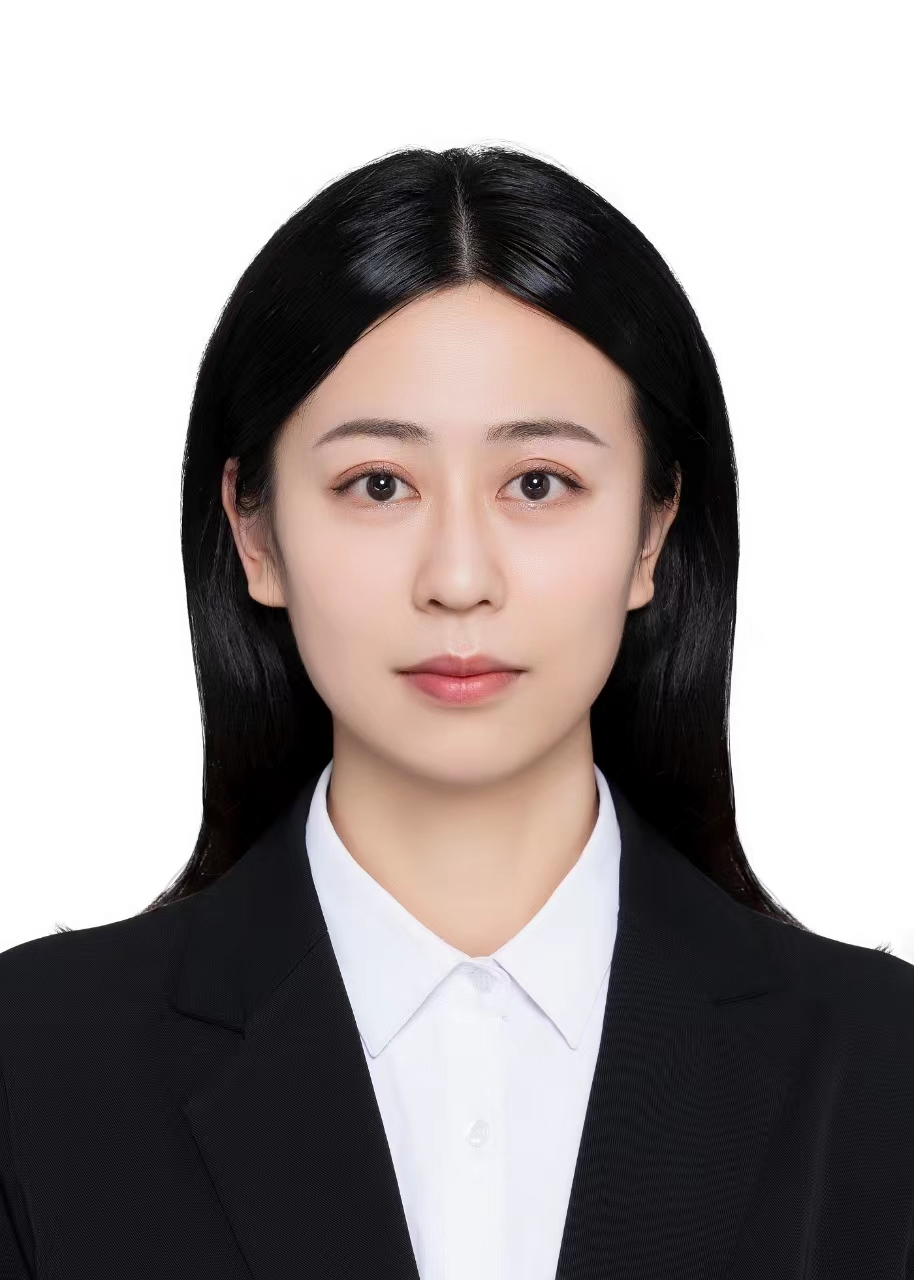}}]
{Chaoqun Wang} is now an associate professor in South China Normal University. She received the B.S. degree in 2018 and the Ph.D. degree in 2023 from University of Science and Technology of China. She took research internship at Microsoft Research Asia in 2017 and 2021. Her major research interests include computer vision problems related to recognition and retrieval.
\end{IEEEbiography}
\vspace{-50mm}

\begin{IEEEbiography}
[{\includegraphics[width=1in,clip,keepaspectratio]{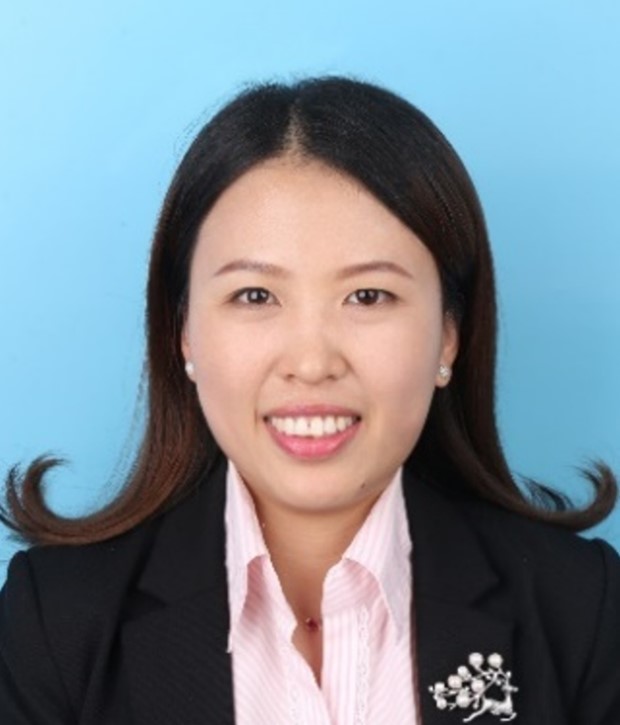}}]
{Xuejin Chen} received the B.S. and Ph.D. degrees in electronic circuits and systems from the University of Science and Technology of China, Hefei, China, in 2003 and 2008, respectively. From 2008 to 2010, she conducted research as a postdoctoral scholar with the Department of Computer Science, Yale University, New Haven, CT, USA. She is currently a Professor with the School of Information Science and Technology, University of Science and Technology of China. Her research interests include 3D modeling, geometry processing, and content creation. She has authored or co-authored over 70 articles in these areas. Dr. Chen was one of the recipients of the Honorable Mention Awards of Computational Visual Media Journal in 2019.
\end{IEEEbiography}

\end{document}